\documentclass[runningheads]{llncs}
\usepackage{graphicx}
\usepackage{amsmath}
\usepackage{amssymb}
\usepackage{bm}
\usepackage{algorithm}
\usepackage[noend]{algorithmic}

\usepackage{verbatim}

\begin{document}
\title{Benchmarking Multivariate Time Series Classification Algorithms}
\author{Alejandro Pasos Ruiz, Michael Flynn and Anthony Bagnall}
\authorrunning{Pasos Ruiz et al.}
\institute{
School of Computing Sciences, University of East Anglia, Norwich, 
}
\maketitle
\begin{abstract}
    Time Series Classification (TSC) involved building predictive models for a discrete target variable from ordered, real valued, attributes. Over recent years, a new set of TSC algorithms have been developed which have made significant improvement over the previous state of the art. The main focus has been on univariate TSC, i.e. the problem where each case has a single series and a class label. In reality, it is more common to encounter multivariate TSC (MTSC) problems where multiple series are associated with a single label. Despite this, much less consideration has been given to MTSC than the univariate case. The UEA archive of 30 MTSC problems released in 2018 has made comparison of algorithms easier. We review recently proposed bespoke MTSC algorithms based on deep learning, shapelets and bag of words approaches. The simplest approach to MTSC is to ensemble univariate classifiers over the multivariate dimensions. We compare the bespoke algorithms to these dimension independent approaches on the 26 of the 30  MTSC archive problems where the data are all of equal length. We demonstrate that the independent ensemble of HIVE-COTE classifiers is the most accurate, but that, unlike with univariate classification, dynamic time warping is still competitive at MTSC.

\end{abstract}
\section{Introduction}

 Time series classification (TSC) is a form of machine learning where the features of the input vector are real valued and ordered. This scenario adds a layer of complexity to the problem, as important characteristics of the data can be missed by traditional algorithms.
 Over recent years, a new set of TSC algorithms have been developed which have made significant improvement over the previous state of the art~\cite{bagnall17bakeoff}.

 The main focus has been on univariate TSC, i.e. the problem where each case has a single series and a class label. In reality, it is more common to encounter multivariate TSC (MTSC) problems where multiple series are associated with a single label. Human activity recognition, diagnosis based on ECG, EEG and MEG and systems monitoring problems are all inherently multivariate. Despite this, much less consideration has been given to MTSC than the univariate case. The UCR archive has provided a valuable resource for univariate TSC, and its existence may explain the growth of algorithm development for this task. Until recently, there were few resources for MTSC. An archive of 30 MTSC problems released in 2018~\cite{bagnall18mtsc} has made comparison of algorithms easier and will we hope spur further research in this field. We compare recently proposed bespoke MTSC algorithms to simple adaptations of univariate approaches on the 26 equal length problems in the UEA MTSC archive.

 We compare twelve algorithms, including recent deep learning approaches, bag of words based algorithms and shapelet based classifiers. We find that dynamic time warping (DTW) is still hard to beat in MTSC, but that two algorithms achieve a significant improvement in accuracy of this benchmark.


\section{Background}
\label{sec:background}

In univariate time series classification, an instance is a pair
$\{\bm{x}, y\}$ with $m$ observations $(x_1,\ldots, x_m)$ (the time series) and discrete class variable $y$ with $c$ possible values. A classifier is a function or mapping from the space of possible inputs to a probability distribution over the class variable values. In MTSC, the input is a list of vectors over $d$ dimensions and $m$ observations, $\bm{x}=<\bm{x_1}, \ldots \bm{x_d}>$, where $\bm{x_k}=(x_{1,k},x_{2,k},\ldots,x_{m,k})$. We denote the $j^{th}$ observation of the $i^{th}$ case of dimension $k$ as the scalar $x_{i,j,k}$.

The core additional complexity for MTSC is that discriminatory features may be in the interactions between dimensions, not just in the autocorrelation within a series, or that the sheer volume of data may obscure discriminatory features. Algorithms for MTSC can be categorised in the same way as algorithms for univariate TSC on whether they are based on: distance measures; shapelets; histograms over a dictionary; or deep learning/neural networks. Distance based approaches are mainly based on dynamic time warping (DTW). Three dynamic time warping approaches proposed in~\cite{shokoohi17generalizing} are described in Section~\ref{sec:distance}.
DTW approaches are a popular benchmark for TSC. Another obvious benchmark is to adapt univariate algorithms to the multivariate case by ensembling over dimensions. We elaborate on this in Section~\ref{sec:hive-cote}.

The shapelet based approach most geared towards MTSC involves forests of decision trees~\cite{karlsson16generalized}. Details of the generalized random shapelet forest~\cite{karlsson15forests} are given in Section~\ref{sec:shapelets}. The most effective dictionary (bag-of-patterns) algorithm for MTSC is the Word extraction for time series classification (WEASEL) with a Multivariate Unsupervised Symbols and dErivatives (MUSE)~\cite{schafer17weasel+muse}. We refer to this combination WEASEL+MUSE as simply MUSE and describe it it in Section~\ref{sec:weasel}. A range of deep learning approaches have been developed for TSC. Two variants meant specifically for MTSC, the  Multivariate  Long  Short  Term  Memory  Fully  Convolutional Network  (MLCN)~\cite{karim19lstm} and the Time Series Attentional Prototype Network (TapNet)~\cite{zhang20tapnet} are described in Section~\ref{sec:mlstmfcn}.


\subsection{Dynamic Time Warping}
\label{sec:distance}

 One of the most popular approaches for TSC is to use a 1-Nearest neighbourhood classifier in conjunction with a bespoke distance function that compensates for possible confounding offset by allowing some realignment of the series.  Dynamic time warping (DTW) is the most popular distance function for this purpose. In DTW, the distance between series $\textbf{a}=(a_1,a_2,..., a_m)$ and $\textbf{b}=(b_1,b_2,..., b_m)$ is calculated following these steps:

\begin{enumerate}
    \item $M$ is a $m \times m$ matrix where $M_{i,j}=(a_i-b_j)^2$
    \item A warping path $P=((e_1,f_1),(e_2,f_2),...,(e_s,f_s))$ is a contiguous set of matrix indexes from $M$, subject to the following constraints
    \begin{itemize}
        \item $(e_1,f_1)=(1,1)$
        \item $(e_s,f_s)=(m,m)$
        \item $0 \leq e_{i+1} - e_i \leq 1$ for all $i<m$
        \item $0 \leq f_{i+1} - f_i \leq 1$ for all $i<m$

    \end{itemize}
    \item Let $p_i=M_{e_i,f_i}$, be the distance for a path is $D_p=\sum_{i=1}^m p_i$
    \item There are many warping paths but we are interested in the one that minimizes the accumulative distance $P^*=\min_{p \in P} D_p(a,b)$
    \item The optimal distance is obtained by solving the following recurrence relation
    \begin{equation}
      DTW(i,j)=M_{i,j}+ min\begin{cases}
        DTW(i-1, j).\\
        DTW(i, j-1).\\
        DTW(i-1, j-1).
      \end{cases}
    \end{equation}
    and the final distance is $DTW(m,m)$.
\end{enumerate}
There are several improvements to DTW to make it faster, such as, for example, adding a parameter $r$ that limits deviation from the diagonal. Our interest lies primarily in how best to use DTW for MTSC. There are two obvious strategies for using DTW for multivariate problems, defined as the independent and dependent approaches.

\subsubsection{Independent Warping, ($DTW_I$)}

The independent strategy treats each dimension independently, has a different pointwise distance matrix $M$ for each dimension, then sums the resulting DTW distances.

\begin{equation}
    DTW_I(\textbf{a},\textbf{b}) = \sum_{k=1}^d DTW(\bm{a_k},\bm{b_k})
\end{equation}

 \subsubsection{Dependent ($DTW_D$)}

 Respectively, this assumes some relation among the series on the multivariate time series. For handling this case, the matrix $M_{i,j}$ is redefined not as the distance between 2 points on a single series but as the euclidean distance between the 2 vectors that represent all the series.

 \begin{equation}
    M_{i,j}=\sum_{k=1}^d (a_{i,k}-b_{j,k})^2
\end{equation}

 Then, the DTW function or euclidean distance is calculated which leads to distance time wrapping dependent($DTW_D$) or euclidean distance dependent($ED_D$)

 \subsubsection{Adaptive ($DTW_A$)}

  Shokoohi-Yekta et al.  \cite{shokoohi2017generalizing}  discussed when a problem is independent or dependent. This discussion led to create the adaptive case which uses a combination of the previous two and tries to define when it should be used. This method uses the dependent or independent distance depending on a threshold. The threshold is calculated on the training phase using cross validation on training data and verifying which distance does better.


\subsection{Ensembles of Univariate Classifiers}
\label{sec:hive-cote}

One of the most straightforward techniques to adapt TSC algorithms to multivariate is to consider independence over dimensions and ignore relations among them. This approach is a good baseline for assessing bespoke MTSC. One of the most accurate approaches to univariate TSC is the Hierarchical Vote Collective of Transformation-based Ensembles (HIVE-COTE). The latest version, HIVE-COTE v1.0 (referred to as simply HIVE-COTE)~\cite{bagnall20report3}, combines
Shapelet Transform Classifier (STC)~\cite{hills14shapelet}; Time Series Forest (TSF)~\cite{deng13forest}; Bag of Symbolic-Fourier Approximation Symbols (BOSS)~\cite{schafer15boss} and Random Interval Spectral Ensemble (RISE)\cite{lines18hive} using a weighted probabilistic ensemble (CAWPE) \cite{large19cawpe}.
The simplest way to build a multivariate HIVE-COTE is to build each component as an independent ensemble, then to combine the components in the usual way. To clarify, each component builds a separate classifier on every dimension, then combines the predictions from each dimension to produce a single probability distribution for each component. In addition to HIVE-COTE, each independent classifier (STC, TSF, BOSS, RISE) were run alone in order to show how each algorithm type performs.

\subsection{Generalized Random Shapelet Forest (gRFS)}
\label{sec:shapelets}
Shapelets can be defined as time series sub sequences and since their inception have commanded a lot of attention. One of their biggest draws is the high interprebility, allowing practitioners to identify class defining features. However, the train time associated with full enumeration is prohibitive and has been shown to grow exponentially with respect to series length~\cite{rakthanmanon13fastshapelets}.

Karlsson \textit{et al.}~\cite{karlsson16generalized} propose a forest approach deliberately designed to be used in conjunction with multivariate datasets. The proposed approach is inspired by Breimans work on bagging~\cite{breiman1996bagging} and the well known RandomForest~\cite{breiman2001random} in which randomisation is used to reduce train time and increase variability throughout an ensemble.

\begin{algorithm}
    \caption{Random Shapelet Forest($\boldsymbol{Z}, p, l, u, r$)}
    \label{algo:gsf}
    \begin{algorithmic}[1]
        \REQUIRE The training set, $\boldsymbol{Z}$, the number of trees, $p$, the lower and upper shapelet length, $l$ \& $u$, the number of shapelets, $r$.
        \ENSURE An ensemble of generalized shapelet trees, $\boldsymbol{R}$ $=$ {$\boldsymbol{ST}_1$ \dots  $\boldsymbol{ST}_p$}
        \FOR{$i \gets 1$ to $p$}
            \STATE $\boldsymbol{I}_i \gets$ sample($\boldsymbol{Z}$)
            \STATE $\boldsymbol{ST}_i$ $\gets$ randomShapeletTree($\boldsymbol{Z}_{\boldsymbol{I}i}$, $l$, $u$, $r$)
            \STATE $\boldsymbol{R} \gets \boldsymbol{R} \cup \boldsymbol{ST}_i$
        \ENDFOR
        \RETURN $\boldsymbol{R}$
    \end{algorithmic}
\end{algorithm}

The Generalised Random Forest is an ensemble of weak learners in which $p$ generalized trees are grown, illustrated in algorithms~\ref{algo:gsf}~\&~\ref{algo:rst}. In order to introduce variability amongst the constituent classifiers a bagging approach is employed. At each node in a generalised tree a dimension, $k$, is randomly selected to proceed with. From this $r$ shapelets are selected. Each shapelet, \textbf{S$_{k,i}^{j,l}$}, has a randomly selected length, \textbf{S$^l$}, such that $l \leq$  \textbf{S$^l$} $\leq u$. Where $l$ and $u$ are parameters for the minimum and maximum allowable length and a randomly selected start position, \textbf{S$^j$}, defined as \textbf{S$^j$} $= rand(1, m -$ \textbf{S$^l$}$)$, where $m$ is the number of attributes in a instance. The shapelet selected at each node corresponds to that which produces the most favourable split. The measure of goodness selected to asses the performance of a split is entropy. Each tree is grown until the number of instances in each leaf node is less than 3.

\begin{algorithm}
    \caption{Random Shapelet Tree($\boldsymbol{Z}, l, u, r$)}
    \label{algo:rst}
    \begin{algorithmic}[1]
        \REQUIRE The training set, $\boldsymbol{Z}$, the lower and upper shapelet length, $l$ \& $u$, the number of shapelets, $r$.
        \ENSURE A random shapelet tree, $\boldsymbol{ST}$
        \IF{isTerminal($\boldsymbol{Z}$)}
            \RETURN makeLeaf($\boldsymbol{Z}$)
        \ENDIF
        \FOR{$i \gets 1$ to $r$}
            \STATE $\boldsymbol{S} \gets \boldsymbol{S}$ $\cup$ sampleShapelet($\boldsymbol{Z}$, $l$, $u$, rand($l$, $u$))
        \ENDFOR
        \STATE [$t$, $S$, $k$] $\gets$ bestSplit($\boldsymbol{Z}$, $y$, $\boldsymbol{S}$)
        \STATE [$\boldsymbol{Z}_L$, $\boldsymbol{Z}_R$] $\gets$ distribute($\boldsymbol{Z}$, $S$, $t$, $k$)
        \STATE $ST_L \gets$ randomShapeletTree($\boldsymbol{Z}_L$, $l$, $u$, $r$)
        \STATE $ST_R \gets$ randomShapeletTree($\boldsymbol{Z}_R$, $l$, $u$, $r$)
        \RETURN [[$t$, $S$, $k$, $\boldsymbol{ST}_L$], [$t$, $S$, $k$, $\boldsymbol{ST}_L$]]
    \end{algorithmic}
\end{algorithm}

\subsection{WEASEL+MUSE}
Word extraction for time series classification (WEASEL)~\cite{schafer17fast} was extended to include a Multivariate Unsupervised Symbols and dErivatives (MUSE) stage for MTSC. The classifier finds discriminatory words from each dimension using Algorithm~\ref{sec:weasel} before performing a feature selection with a $\chi^2$ test. Final classification is with a logistic regressor.  For simplicity, we will refer to this algorithm as just MUSE forthwith.
\label{sec:weasel}
        \begin{algorithm}[htb]
    	\caption{WEASEL+MUSE(A list of $n$ cases of length $m$ with dimension $d$, ${\bf T}=({\bf X,y})$)}
    	\label{alg:weasel}
    	\begin{algorithmic}[1]
    	\REQUIRE the word length $l$, the alphabet size $\alpha$, the maximal window length $w_{max}$, mean normalisation parameter $p$
    		\STATE Let ${\bf H}$ be the histogram ${\bf h}$
    		\STATE Let ${\bf B}$ be a matrix of $l$ by $\alpha$ breakpoints found by MCB using information gain binning
    		\FOR {$i \leftarrow  1$ to $n$}
        		\FOR {$k \leftarrow  1$ to $d$}
        			\FOR {$w \leftarrow  2$ to $w_{max}$}
        				\FOR {$j \leftarrow 1$ to $m-w+1$}
        					\STATE ${\bf o}\leftarrow x_{i,j,k} \ldots x_{i,j+w-1,k}$
        					\STATE ${\bf q} \leftarrow$ DFT($o, w, p$) \COMMENT{ {\em {\bf q} is a vector of 	the complex DFT coefficients}}
        					\STATE ${\bf q'} \leftarrow$ ANOVA-F($q, l, y$) \COMMENT{ {\em use only the {\bf l} most discriminative ones}}					
        					\STATE ${\bf r} \leftarrow$ SFAlookup(${\bf q', B}$)
        					\STATE $pos \leftarrow $index(${\bf w, r}$)
        					\STATE ${h}_{i,pos} \leftarrow {h}_{i,pos} + 1$
        				\ENDFOR					
        			\ENDFOR
        		\ENDFOR		
    		\ENDFOR
    		\STATE $h \leftarrow \chi^2(h, y)$ \COMMENT{ {\em feature selection using the chi-squared test} }
    		\STATE fitLogistic($h, y$)
    	\end{algorithmic}
        \end{algorithm}

\subsection{Deep Learning}

Recently, most approaches to MTSC have used neural networks, and in particular convolutional neural networks. TapNet~\cite{zhang20tapnet} uses an attentional prototype network to learn the latent features.  A previously proposed Long Short Term Memory Convolutional Network for time series classification (LSTM-FCN) ~\cite{karim2017lstm} is adapted for the multivariate case and the fully convolutional blocks augmented with squeeze-and-excitation blocks to form MLSTM-FCN~\cite{karim19lstm}. For the sake of space, we shall refer to MLSTM-FCN as MLCN.


\subsubsection{The Multivariate Long Short Term Memory Fully Convolutional Network (MLCN).}
\label{sec:mlstmfcn}

\begin{figure}[t]
    \centering
    \makebox[\textwidth][c]{\includegraphics[width=\textwidth, trim={.2cm .2cm .2cm .2cm},clip]{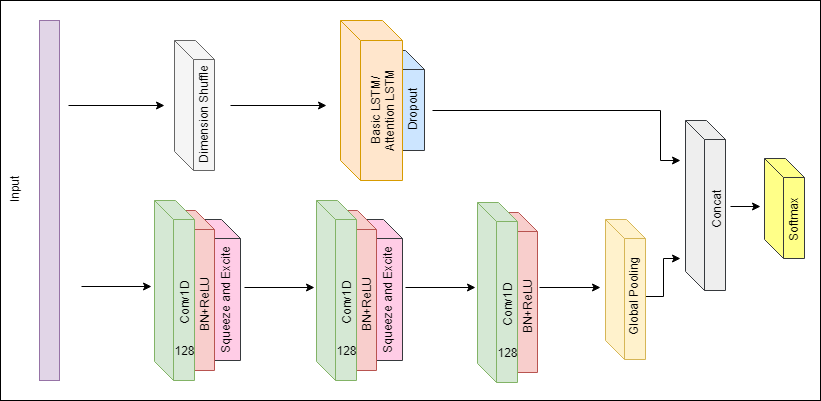}}
    \caption{MLSTM-FCN architecture, figure from~\cite{karim19lstm}.}
    \label{fig:mlstmfcn}
\end{figure}

MLCN~\cite{karim19lstm} builds on the architecture originally used for univariate time series classification~\cite{karim2017lstm}. Fundamentally, it is a combination of long short term memory (LSTM)~\cite{hochreiter1997long} and one-dimensional fully convolutional networks (FCN)~\cite{wang17fcn} joined by a concatenation layer, followed by a shared dense layer for predictions. The architecture is summarised in Figure~\ref{fig:mlstmfcn}. LSTMs are a form of recurrent neural network able to learn temporal dependencies in sequences, while the FCNs act as a learned feature extraction process. Combining these and learning from both simultaneously provides good theoretical coverage of the numerous different types of problems that may be faced in time series classification. The extensions to the multivariate case comes with the addition of squeeze-and-excitation blocks for the first two convolutional blocks, which model and adaptively scale dependencies between feature maps at each stage.

~\cite{karim19lstm} describes two forms of the network, with and without attention mechanisms for the LSTM layer. These were found to be not significantly different from one another over 35 datasets, with wins and losses across different datasets. We take the MLSTM-FCN (without the attention mechanism) for comparison to reduce comparative bloat in analysis, and for ease and speed of reproduction.

In the original evaluations, network hyperparameters are fixed across all datasets with the exception of the number of LSTM cells, which is optimised on the train data in the range $\{8, 64, 128\}$. We maintain all hyperparameter settings from the original work, however for the comparisons presented here, we have fixed the number of LSTM cells to 64 for all datasets. 

\subsubsection{Time Series Attentional Prototype Network (TapNet).}
\label{sec:tapnet}
A novel approach aimed at tackling problems in the multivariate domain, the TapNet architecture draws on the strengths of both traditional and deep learning approaches. Zang \textit{et al.}~\cite{zhang20tapnet} note that deep learning approaches excel at learning low dimensional features without the need for embedded domain knowledge whereas traditional approaches such as 1NN-DTW work well on comparatively small datasets. TapNet combines these advantages to produce a network architecture that can be broken down into 3 distinct modules: Random Dimension Permutation, Multivariate Time Series Encoding and Attentional Prototype Learning.
\begin{figure}[t]
    \centering
    \makebox[\textwidth][c]{\includegraphics[width=1.25\textwidth, trim={1cm 0cm 1cm 0cm},clip]{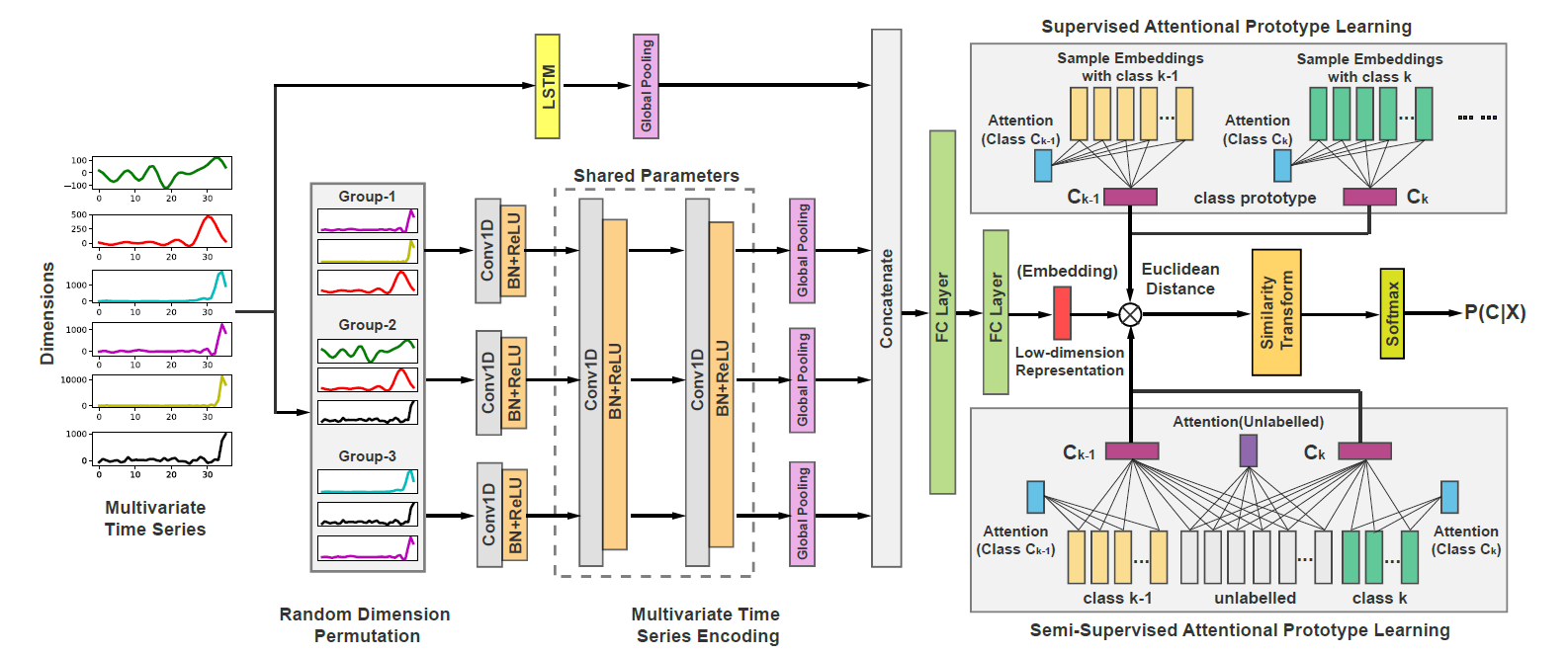}}
    \caption{TapNet architecture, figure from~\cite{zhang20tapnet}.}
    \label{fig:tapnet}
\end{figure}

Random Dimension Permutation is used to produce $g$ groups of randomly selected dimensions with the intention of increasing the likelihood of learning how combinations of dimension values effect class value. The group size is defined as $\varphi = \lfloor \frac{m \cdot \alpha}{g} \rfloor$, where $\alpha$ is the scale factor, controlling the number of dimensions used over $m$, where $m$ is the number of dimensions. This process is illustrated in figure~\ref{fig:tapnet} where the six input dimensions are reorganised into three groups of three. Experimentation exploring the effect of this module found that in 22 out of 33 datasets in the UEA multivariate archive the accuracy was increased. However, it is unclear whether it has a significant effect or whether the effect on accuracy is a function of dataset characteristics.

Encoding in the TapNet architecture is undertaken in $g + 1$ stages before the output features are concatenated and passed through two fully connected layers. Each group produced in the dimension permutation module is passed through three sets of one-dimensional convolutional layers followed by batch normalisation, Leaky Rectified Linear Units and finally a global pooling layer. For the first of these three sets the weights and bias are distinct for each group. Additionally to the group encoding process the raw data is passed through an LSTM and global pooling layer. The output from each of the global pooling layers are then concatenated before being passed through two fully connected layers. This process results in a low-dimensional feature representation of the original series. The default filter values for the convolution layers are set as 256, 256 and 128 whilst the default kernel values are 5, 8 and 3. The default value for the LSTM layer is 128. It is intended that interaction between dimensions can be learned more effectively by the Random Dimension Permutation process before the encoding is then combined, producing features aligned with a datasets dimensions. Furthermore, the inclusion of the LSTM layer is intended to learn longitudinal features.

Finally, for each class a prototype candidate is produced. The objective of the candidate production is to minimise the distance to all members of the class which the prototype is produced for whilst maximising the distance between the prototypes. Probability of class membership is then assigned to test instances as a function of their proximity to each class prototype. In this case the similarity is measured by way of euclidean distance.

\section{Data}
 \label{sec:data}

\begin{table}[t!]
\begin{center}
\caption{Summary of the 26 UEA datasets used in experimentation.}
\begin{tabular}{|l|l|c|c|c|c|c|}
\hline
   &  Name & Train size & Test size & Num Series & Series length & Classes  \\
\hline
AWR     &     ArticularyWordRecognition & 275 & 300 & 9 & 144 & 25  \\
AF      &     AtrialFibrillation & 15 & 15 & 2 & 640 & 3 \\
BM      &     BasicMotions & 40 & 40 & 6 & 100 & 4 \\
CR      &    Cricket & 108 & 72 & 6 & 1197 & 12 \\
DDG     &    DuckDuckGeese & 50 & 50 & 1345 & 270 & 5 \\
EW      &    EigenWorms & 128 & 131 & 6 & 17984 & 5 \\
EP      &    Epilepsy & 137 & 138 & 3 & 206 & 4 \\
EC      &    EthanolConcentration & 261 & 263 & 3 & 1751 & 4 \\
ER      &    ERing & 30 & 270 & 4 & 65 & 6 \\
FD      &    FaceDetection & 5890 & 3524 & 144 & 62 & 2 \\
FM      &    FingerMovements & 316 & 100 & 28 & 50 & 2 \\
HMD     &    HandMovementDirection & 160 & 74 & 10 & 400 & 4 \\
HW      &    Handwriting & 150 & 850 & 3 & 152 & 26 \\
HB      &    Heartbeat & 204 & 205 & 61 & 405 & 2 \\
LIB     &    Libras & 180 & 180 & 2 & 45 & 15 \\
LSST    &    LSST & 2459 & 2466 & 6 & 36 & 14 \\
MI      &    MotorImagery & 278 & 100 & 64 & 3000 & 2 \\
NATO    &    NATOPS & 180 & 180 & 24 & 51 & 6 \\
PD      &    PenDigits & 7494 & 3498 & 2 & 8 & 10 \\
PEMS    &    PEMS-SF & 267 & 173 & 963 & 144 & 7 \\
PS      &    PhonemeSpectra& 3315 & 3353 & 11 & 217 & 39 \\
RS      &    RacketSports & 151 & 152 & 6 & 30 & 4   \\
SRS1  &      SelfRegulationSCP1 & 268 & 293 & 6 & 896 & 2 \\
SRS2  &      SelfRegulationSCP2 & 200 & 180 & 7 & 1152 & 2 \\
SWJ  &       StandWalkJump & 12 & 15 & 4 & 2500 & 3 \\
UW  &     UWaveGestureLibrary & 120 & 320 & 3 & 315 & 8 \\
\hline
\end{tabular}
\label{table:data}
\end{center}
\end{table}
The UEA MTSC archive~\cite{bagnall18mtsc} released in 2018 contained multivariate datasets, of which four are not all equal length. To focus on classification rather than preprocessing, we restrict our attention to the 26 equal length series. The main characteristics of each one is summarized in Table~\ref{table:data}. Details can be found on the associated website\footnote{https://www.timeseriesclassification.com/}.

\section{Results}
 \label{sec:results}

Our experiments are conducted with a range of software tools. All variants of DTW and HIVE-COTE components are run using the tsml toolkit\footnote{https://github.com/uea-machine-learning/tsml}. They are also available in the the scikit learn like aeon toolkit\footnote{https://github.com/aeon-toolkit/aeon}. MUSE and TapNet are also available in aeon.

Each dataset is provided with a default split of train and test datasets. Table~\ref{table:data} shows the number of train and test instances for each of the datasets used. Each experiment is limited to a maximum seven day execution time and 500 GB memory. TapNet completed on  23 datasets, crashing with memory errors for PhonemeSpectra, EigenWorms and MotorImagery. We ran gRSF with default parameters on all datasets without problems. However, tuning with the recommended parameter ranges~\cite{karlsson15forests} proved infeasible. Only 9 of the 26 experiments completed in 7 days. The bottleneck for MUSE is memory. Four of the  26 problems, DuckDuckGeese, MotorImagery, PEMS-SF and PhonemeSpectra, could not complete with 500GB memory. All classifiers completed 21 problems, 10 classifiers completed all 26.

\subsection{Comparison of Classifiers}
\begin{figure}[t]
	\centering
    \includegraphics[width=\linewidth,trim={0cm 4cm 0cm 5cm},clip]{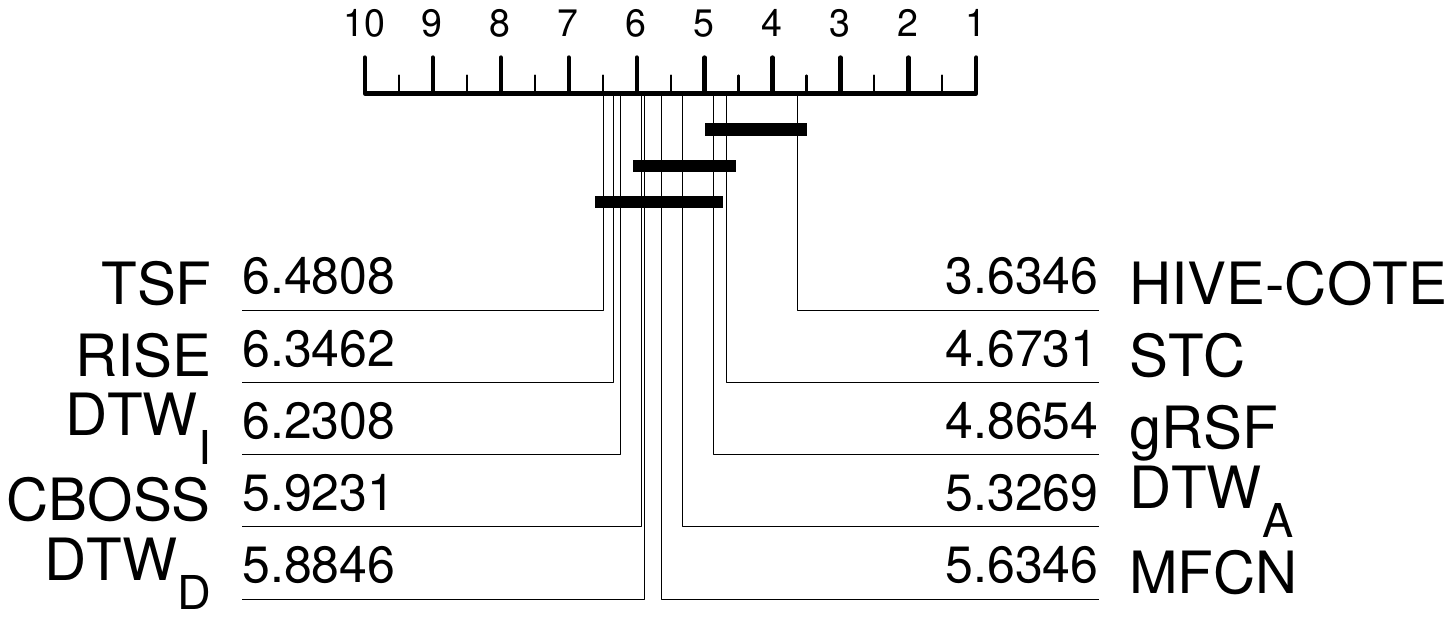}
    \caption{Critical difference diagram for ten multivariate time series classifiers on 26 UEA MTSC time series classification problems. Full results are available on the accompanying website.}
    \label{fig:complete}
\end{figure}

Figure~\ref{fig:complete} shows the critical difference diagram for the ten classifiers which completed on the default train/test split. The number associated with each classifier is the average rank over all 26 problems (lower is better). Solid bars represent cliques. There is no significant difference of between classifiers in the same clique. This is tested using a pairwise a pairwise Wilcoxon signed rank tests and the Holm correction to form cliques.
The top clique consists of HIVE-COTE, STC and gRSF. The second clique contains STC, gRSF, DTW$_A$, MLCN, DTW$_D$, DTW$_D$ and CBOSS. The third clique is all classifiers except STC and HIVE-COTE. Overall, only one classifier, HIVE-COTE, is significantly better than the standard benchmark, DTW$_D$.

\begin{figure}[t]
	\centering
    \includegraphics[width=\linewidth,trim={0cm 2cm 0cm 5cm},clip]{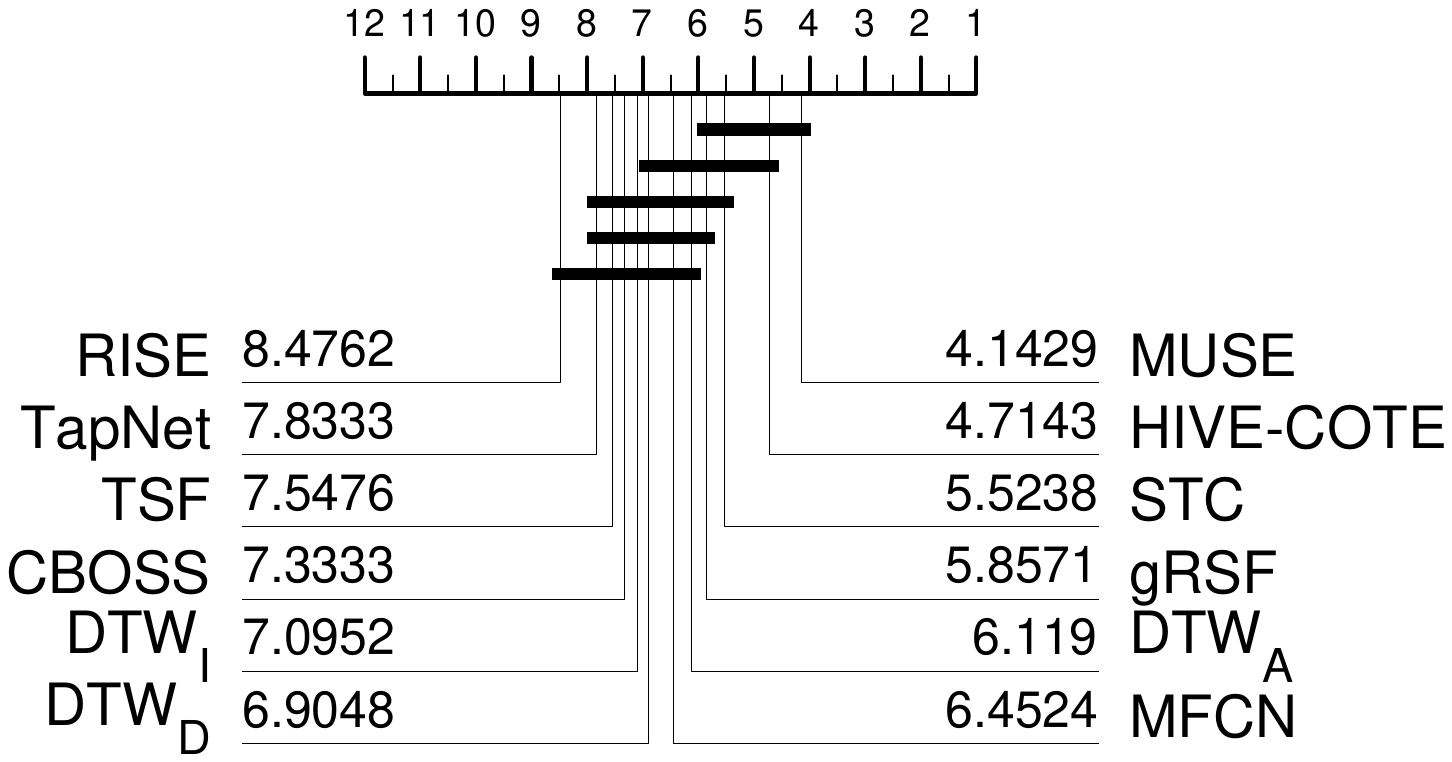}
    \caption{Critical difference diagram for 12 multivariate time series classifiers on 21 UEA MTSC time series classification problems. Full results are available on the accompanying website.}
    \label{fig:all}
\end{figure}

Figure~\ref{fig:all} shows the critical difference diagram for twelve classifiers on the 21 data sets that all algorithms completed within our constraints. MUSE joins the top clique, and, like HIVE-COTE, is significantly more accurate than DTW$_D$. The two deep learning approaches, TapNet and MLCN, do not perform as well as the shapelet algorithms and are no better than any of the three DTW algorithms. RISE is the worse performing algorithm overall, but it does relatively well on certain problems.

All of these results are obtained without normalisation. We have run the experiments with each dimension independently normalised. It made no significant difference to the DTW classifiers. All three were worse on average, although not significantly so. We also built HIVE-COTE and its components on normalised series, and found no significant difference.

\begin{figure}[thb]
	\centering
    \includegraphics[width=\linewidth]{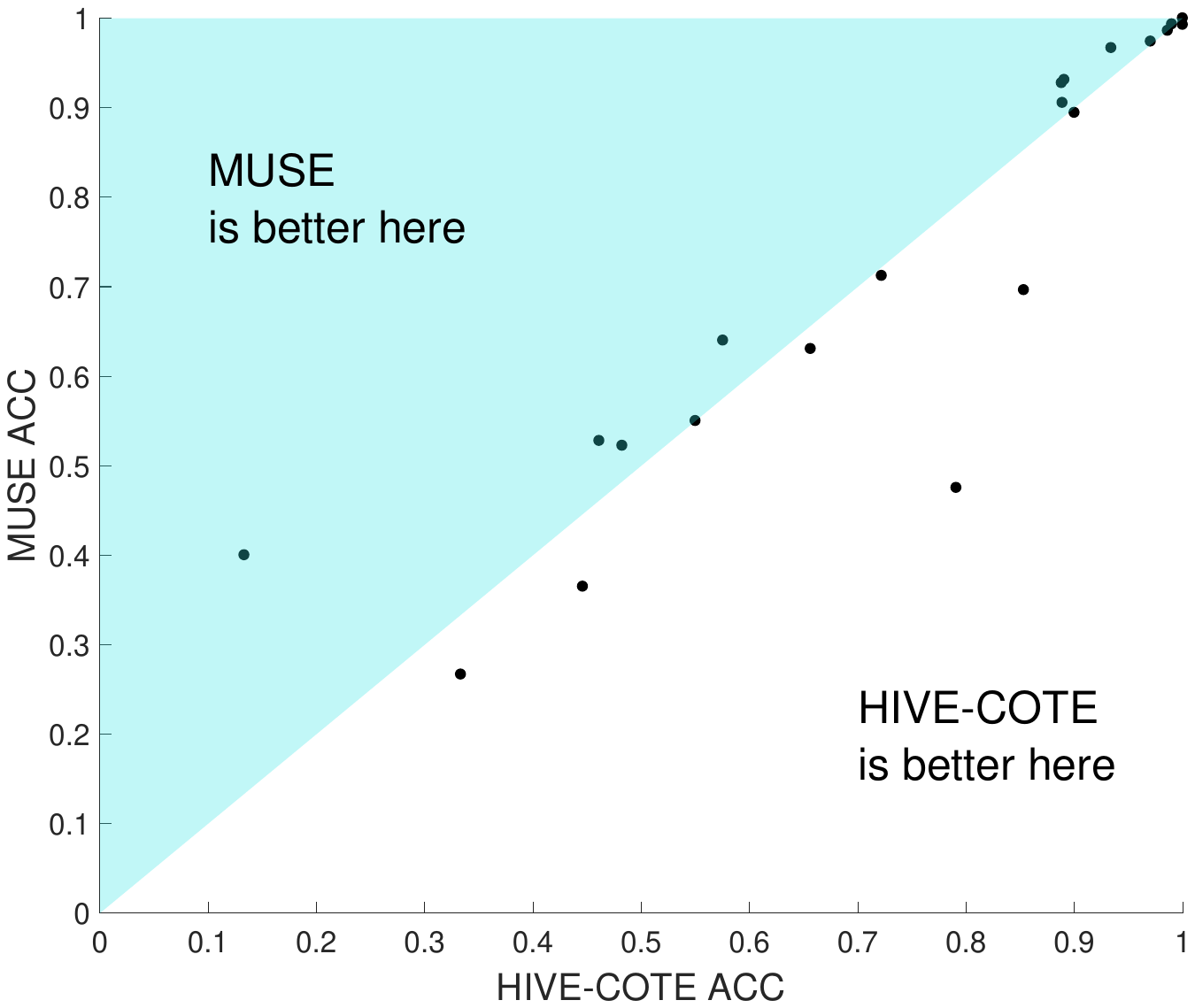}
    \caption{MUSE vs HIVE COTE on 21 UEA MTSC problems.}
    \label{fig:muse-hc}
\end{figure}

Figure~\ref{fig:muse-hc} shows the scatter plot of test accuracies for HIVE-COTE against the standard benchmark, DTW$_D$. HIVE-COTE is more accurate on 19 and less accurate on 7. The median difference in accuracy is 2.46\%.
\begin{figure}[thb]
	\centering
    \includegraphics[width=\linewidth]{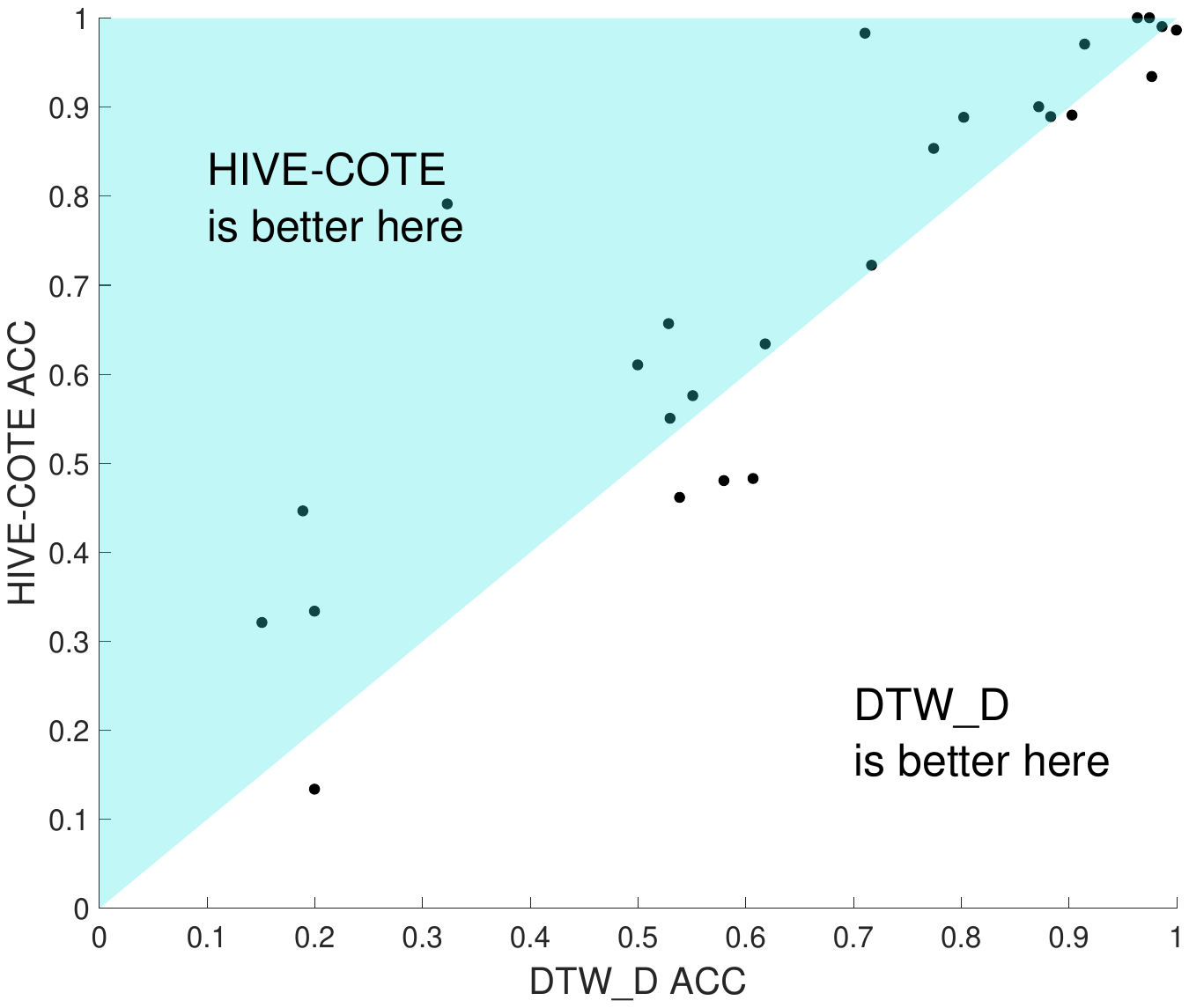}
    \caption{DTW$_D$ vs HIVE COTE on 26 UEA MTSC problems.}
    \label{fig:muse-dtw}
\end{figure}
Figure~\ref{fig:muse-hc} shows the scatter plot of test accuracies for MUSE vs HIVE-COTE. MUSE wins on 10, HIVE-COTE on 8 and they tie on three. There is wide diversity between the algorithms on seven of the problems. This indicates that HIVE-COTE may benefit from inclusion of inter-dimension relationships.
Run times and memory are hard to compare. The neural networks were run on a GPU, gRSF in python and the HIVE-COTE components were distributed across many CPU, with a model built on each dimension in parallel. Qualitatively, MUSE is relatively fast but very memory intensive. DTW$_A$ is slower than MUSE but faster than HIVE-COTE, and less memory intensive than either of the other two. If HIVE-COTE were run sequentially, it would be much slower than either DTW$_A$ or MUSE, but use an order of memory less than MUSE. HIVE-COTE is also contractable, meaning a maximum run time can be set.

\subsection{Analysis by Problem}

\begin{table}[ht]
\caption{Accuracy of 12 algorithms on the default Train/Test data sets for the UEA MTSC archive.}

\begin{center}
\scalebox{0.85}{
\begin{tabular}{|l|c|c|c|c|c|c|c|c|c|c|c|c|c|}
\hline

Problem  &  Default  &  HC  &  STC  &  gRSF  &  DTW$_A$  &  MLCN  &  DTW$_D$  &  CBOSS  &  DTW$_I$  &  RISE  &  TSF  &  TapNet  &  MUSE \\ \hline
AWR     &  0.04  &  0.99  &  0.99  &  0.983  &  0.987  &  0.957  &  0.987  &  0.99  &  0.98  &  0.963  &  0.953  &  0.957  &  \bf{0.993}\\
AF      & 0.333  &  0.133  &  0.267  &  0.267  &  0.267  &  0.333  &  0.2  &  0.267  &  0.267  &  0.267  &  0.2  &  0.2  &  \bf{0.4}\\
BM      &  0.25  &  \bf{1}  &  0.975  &  \bf{1}  &  \bf{1}  &  0.875  &  0.975  &  \bf{1}  &  \bf{1}  &  \bf{1}  &  \bf{1}  &  \bf{1}  &  \bf{1}\\
CR      &  0.083  &  0.986  &  0.986  &  0.986  &  \bf{1}  &  0.917  &  \bf{1}  &  0.986  &  0.986  &  0.986  &  0.931  &  \bf{1}  &  0.986\\
DDG     &  0.2  &  0.48  &  0.36  &  0.4  &  0.5  &  0.46  &  \bf{0.58}  &  0.38  &  0.48  &  0.46  &  0.22  & \bf{0.58}   &  \\
EW      &  0.42  &  0.634  &  0.779  &  \bf{0.817}  &  0.611  &  0.504  &  0.618  &  0.618  &  0.603  &  \bf{0.817}  &  0.74  &    &  \\
EP      &  0.268  &  \bf{1}  &  0.993  &  0.978  &  0.978  &  0.732  &  0.964  &  \bf{1}  &  0.978  &  \bf{1}  &  0.978  &  0.957  &  0.993\\
EC      &  0.251  &  0.791  &  \bf{0.821}  &  0.346  &  0.316  &  0.373  &  0.323  &  0.361  &  0.304  &  0.487  &  0.445  &  0.308  &  0.475\\
ER      &  0.167  &  0.97  &  0.889  &  0.952  &  0.926  &  0.941  &  0.915  &  0.907  &  0.919  &  0.859  &  0.881  &  0.904  &  \bf{0.974}\\
FD      &  0.5  &  \bf{0.656}  &  0.646  &  0.548  &  0.528  &  0.555  &  0.529  &  0.516  &  0.513  &  0.508  &  0.64  &  0.603  &  0.631\\
FM      &  0.49  &  0.55  &  0.51  &  \bf{0.58}  &  0.51  &  \bf{0.58}  &  0.53  &  0.48  &  0.52  &  0.56  &  0.58  &  0.47  &  0.55\\
HMD     &  0.189  &  0.446  &  0.392  &  0.419  &  0.203  &  \bf{0.527}  &  0.189  &  0.189  &  0.297  &  0.297  &  0.486  &  0.338  &  0.365\\
HW      &  0.038  &  0.482  &  0.288  &  0.375  &  \bf{0.607}  &  0.309  &  \bf{0.607}  &  0.472  &  0.509  &  0.194  &  0.366  &  0.281  &  0.522\\
HB      &  0.722  &  0.722  &  0.722  &  0.761  &  0.693  &  0.38  &  0.717  &  0.722  &  0.659  &  0.732  &  0.741  &  \bf{0.79}  &  0.712\\
LIB     &  0.067  &  \bf{0.9}  &  0.861  &  0.694  &  0.883  &  0.85  &  0.872  &  0.844  &  0.894  &  0.806  &  0.806  &  0.878  &  0.894\\
LSST    &  0.315  &  0.575  &  0.587  &  0.588  &  0.567  &  0.528  &  0.551  &  0.435  &  0.575  &  0.509  &  0.35  &  0.513  &  \bf{0.64}\\
MI      &  0.5  &  \bf{0.61}  &  0.5  &  0.5  &  0.5  &  0.51  &  0.5  &  0.59  &  0.39  &  0.55  & 0.48   &    &  \\
NATO    &  0.167  &  0.889  &  0.872  &  0.844  &  0.883  &  0.9  &  0.883  &  0.861  &  0.85  &  0.839  &  0.8  &  0.811  &  \bf{0.906}\\
PD      &  0.104  &  0.934  &  0.941  &  0.935  &  0.977  &  \bf{0.979}  &  0.977  &  0.908  &  0.939  &  0.832  &  0.892  &  0.856  &  0.967\\
PEMS    &  0.116  &  0.983  &  0.971  &  0.908  &  0.734  &  0.746  &  0.711  &  0.983  &  0.734  &  \bf{0.994}  &  0.983  & 0.78   &  \\
PS      &  0.026  &  \bf{0.321}  &  0.295  &  0.224  &  0.151  &  0.135  &  0.151  &  0.195  &  0.151  &  0.269  &  0.137  &    &\\
RS      &  0.283  &  0.888  &  0.888  &  0.882  &  0.842  &  0.842  &  0.803  &  0.882  &  0.842  &  0.809  &  0.888  &  0.875  &  \bf{0.928}\\
SRS1  &  0.502  &  0.853  &  0.84  &  0.823  &  0.785  &  0.908  &  0.775  &  0.805  &  0.765  &  0.724  &  0.84  &  \bf{0.935}  &  0.696\\
SRS2  &  0.5  &  0.461  &  0.533  &  0.517  &  0.522  &  0.506  &  \bf{0.539}  &  0.489  &  0.533  &  0.494  &  0.483  &  0.483  &  0.528\\
SWJ  &  0.333  &  0.333  &  \bf{0.467}  &  0.333  &  0.333  &  0.4  &  0.2  &  0.333  &  0.333  &  0.267  &  0.333  &  0.133  &  0.267\\
UW  &  0.125  &  0.891  &  0.85  &  0.897  &  0.9  &  0.859  &  0.903  &  0.856  &  0.869  &  0.684  &  0.775  &  0.9  &  \bf{0.931}\\
\hline
Wins  &  0  & 4.444  &  2  & 1.111  &  0.944  &  2.5  &  2.333  &  0.444  &  0.111  &  1.944  &  0.111  &  2.944  &  7.111\\
\hline
\end{tabular}}
\end{center}
\label{tab:acc}
\end{table}

\begin{table}[t!]
\caption{Balanced Accuracy of 12 algorithms on the default Train/Test data sets for the UEA MTSC archive.}

\begin{center}
\scalebox{0.8}{
\begin{tabular}{|l|c|c|c|c|c|c|c|c|c|c|c|c|c|c|}
\hline
Problem  &  Classes  &  Random  &  HC  &   gRSF  &  STC  & DTW$_A$  &  MLCN  &  DTW$_D$  &  CBOSS  &  DTW$_I$  &  RISE  &  TSF  &  TapNet  &  MUSE \\ \hline
AWR    &  25  &  0.04  &  0.99  &  0.983  &  0.99  &  0.987  &  0.957  &  0.987  &  0.98  &  0.99  &  0.953  &  0.963  &  0.957  &  \bf{0.993}\\
AF        &  3  &  0.333  &  0.133  &  0.267  &  0.267  &  0.267  &  0.333  &  0.2  &  0.267  &  0.267  &  0.2  &  0.267  &  0.2  &  \bf{0.4}\\
BM            &  4  &  0.25  &  \bf{1}  &  \bf{1}  &  0.975  &  \bf{1}  &  0.875  &  0.975  &  \bf{1}  &  \bf{1}  &  \bf{1}  &  \bf{1}  &  \bf{1}  &  \bf{1}\\
CR        &  12  &  0.083  &  0.986  &  0.986  &  0.986  &  \bf{1}  &  0.917  &  \bf{1}  &  0.986  &  0.986  &  0.931  &  0.986  &  \bf{1}  &  0.986\\
DDG           &  5  &  0.2  &  0.48  &  0.4  &  0.36  &  0.5  &  0.46  &  \bf{0.58}  &  0.48  &  0.38  &  0.22  &  0.46  &  \bf{0.58}  &  \\
EW            &  5  &  0.2  &  0.484  &  \bf{0.749}  &  0.673  &  0.508  &  0.33  &  0.517  &  0.511  &  0.467  &  0.626  &  0.712  &    &  \\
EP        &  4  &  0.25  &  \bf{1}  &  0.979  &  0.993  &  0.979  &  0.732  &  0.964  &  0.979  &  \bf{1}  &  0.979  &  \bf{1}  &  0.958  &  0.993\\
EC        &  4  &  0.25  &  0.791  &  0.346  &  \bf{0.822}  &  0.316  &  0.373  &  0.323  &  0.304  &  0.361  &  0.445  &  0.487  &  0.309  &  0.476\\
ER        &  6  &  0.167  &  0.97  &  0.952  &  0.889  &  0.926  &  0.941  &  0.915  &  0.919  &  0.907  &  0.881  &  0.859  &  0.904  &  \bf{0.974}\\
FD            &  2  &  0.5  &  \bf{0.656}  &  0.548  &  0.646  &  0.528  &  0.555  &  0.529  &  0.513  &  0.516  &  0.64  &  0.508  &  0.603  &  0.631\\
FM        &  2  &  0.5  &  0.553  &  \bf{0.582}  &  0.512  &  0.509  &  0.58  &  0.53  &  0.519  &  0.489  &  0.581  &  0.562  &  0.471  &  0.551\\
HMD   &  4  &  0.25  &  0.414  &  0.431  &  0.344  &  0.224  &  \bf{0.544}  &  0.206  &  0.292  &  0.21  &  0.481  &  0.312  &  0.338  &  0.362\\
HW    &  26  &  0.038  &  0.478  &  0.372  &  0.285  &  \bf{0.601}  &  0.305  &  \bf{0.601}  &  0.504  &  0.466  &  0.359  &  0.191  &  0.283  &  0.518\\
HB        &  2  &  0.5  &  0.5  &  0.64  &  0.505  &  0.571  &  0.458  &  0.604  &  0.564  &  0.5  &  0.535  &  0.518  &  \bf{0.714}  &  0.515\\
LIB       &  15  &  0.067  &  \bf{0.9}  &  0.694  &  0.861  &  0.883  &  0.85  &  0.872  &  0.894  &  0.844  &  0.806  &  0.806  &  0.878  &  0.894\\
LSST      &  14  &  0.071  &  0.323  &  0.391  &  0.274  &  0.443  &  0.39  &  0.43  &  \bf{0.458}  &  0.153  &  0.161  &  0.265  &  0.386  &  0.435\\
MI        &  2  &  0.5  &  \bf{0.61}  &  0.5  &  0.5  &  0.5  &  0.51  &  0.5  &  0.39  &  0.59  &  0.48  &  0.55  &    &  \\
NATO          &  6  &  0.167  &  0.889  &  0.844  &  0.872  &  0.883  &  0.9  &  0.883  &  0.85  &  0.861  &  0.8  &  0.839  &  0.811  &  \bf{0.906}\\
PD        &  10  &  0.1  &  0.934  &  0.936  &  0.941  &  0.977  &  \bf{0.979}  &  0.977  &  0.939  &  0.908  &  0.892  &  0.831  &  0.855  &  0.967\\
PEMS      &  7  &  0.143  &  0.981  &  0.907  &  0.968  &  0.73  &  0.745  &  0.713  &  0.73  &  0.981  &  0.982  &  \bf{0.994}  & 0.78   &  \\
PS        &  39  &  0.026  &  \bf{0.321}  &  0.224  &  0.295  &  0.151  &  0.135  &  0.151  &  0.151  &  0.195  &  0.137  &  0.269  &    &  \\
RS        &  4  &  0.25  &  0.897  &  0.891  &  0.898  &  0.854  &  0.856  &  0.818  &  0.854  &  0.891  &  0.895  &  0.823  &  0.883  &  \bf{0.933}\\
SRS1  &  2  &  0.5  &  0.854  &  0.823  &  0.84  &  0.786  &  0.908  &  0.775  &  0.765  &  0.806  &  0.84  &  0.724  &  \bf{0.935}  &  0.697\\
SRS2  &  2  &  0.5  &  0.461  &  0.517  &  0.533  &  0.522  &  0.506  &  \bf{0.539}  &  0.533  &  0.489  &  0.483  &  0.494  &  0.483  &  0.528\\
SWJ       &  3  &  0.333  &  0.333  &  0.333  &  \bf{0.467}  &  0.333  &  0.4  &  0.2  &  0.333  &  0.333  &  0.333  &  0.267  &  0.133  &  0.267\\
UW      &  8  &  0.125  &  0.891  &  0.897  &  0.85  &  0.9  &  0.859  &  0.903  &  0.869  &  0.856  &  0.775  &  0.684  &  0.9  &  \bf{0.931}\\ \hline
Wins & &  0  &  4.444  &  2.111  &  2  &  0.944  &  2  & 2.333   &   1.111 & 0.444  &  0.111  & 1.444   &  2.944  & 6.111 \\
\hline
\end{tabular}}
\end{center}
\label{tab:balacc}
\end{table}

\begin{figure}[thb]
	\centering
    \includegraphics[width=\linewidth,trim={0cm 5cm 0cm 5cm},clip]{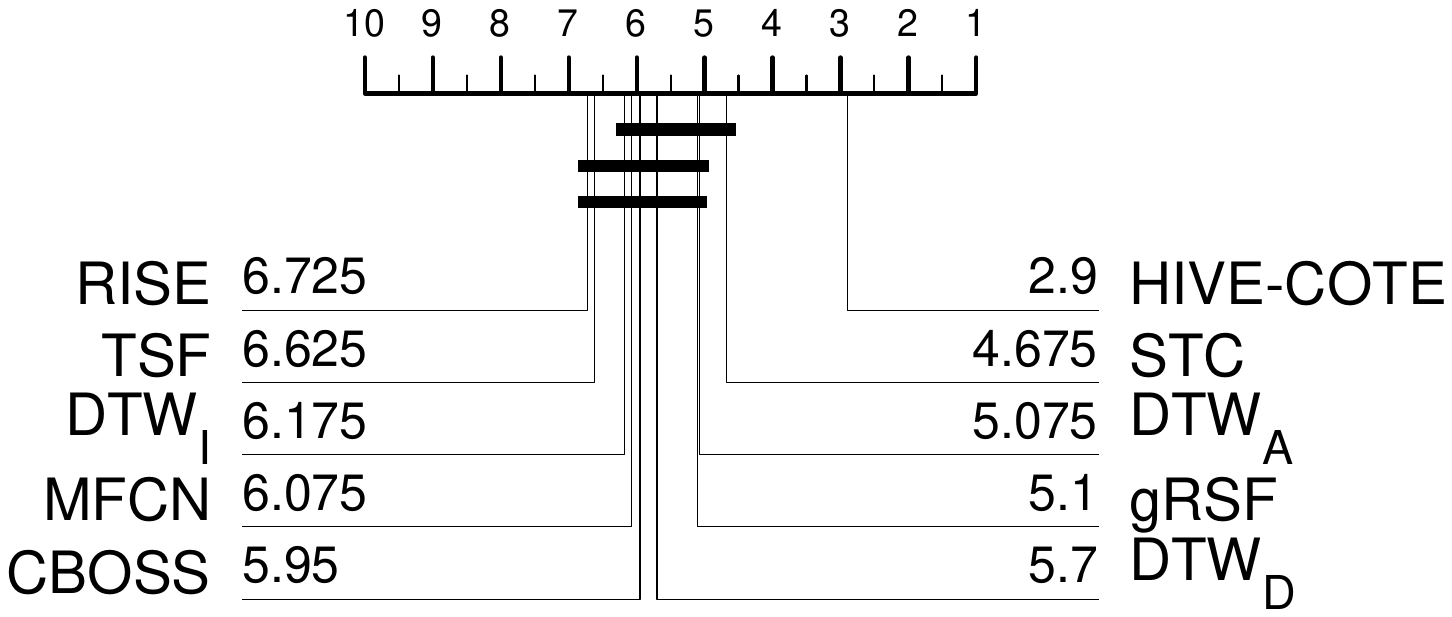}
    \caption{Critical difference diagram for 10 multivariate time series classifiers on 20 UEA MTSC time series classification problems, excluding AF, FM, HB , MI, SRS2 and SWJ.}
    \label{fig:reduced}
\end{figure}

Tables~\ref{tab:acc} and~\ref{tab:balacc} show the test accuracy and the balanced test accuracy respectively. We include the results for predicting the majority class and random guessing so that we can assess how useful these data sets are as benchmarks. If the series contain little or no discriminatory information then using majority class or random guessing will be as good as anything else. We see that with problems AtrialFibrillation (AF), FingerMovements (FM), Heartbeat (HB), MotorImagery (MI), SelfRegulationSCP2 (SRS2) and possibly StandWalkJump (SWJ), few, if any, classifiers do better than predicting the majority class. For problems AF and SWJ, most algorithms do no better than randomly guessing the class. This could have multiple causes. It may be there is no useful information in the attribute space. This is probably the case with SRS2, according to the original data source~\footnote{http://bbci.de/competition/ii/tuebingen\_desc\_ii.html}. If we remove these six problems, we get a clearer picture of improvement gained from ensembling with HIVE-COTE (see figure~\ref{fig:reduced}).
However, it could also be the case that these classifiers are unable to discover the discriminatory information. It could reside in the interaction between dimensions or be swamped by the large number of dimensions.

 \section{Conclusions}

This experimental analysis has demonstrated that MTSC is at an earlier stage of development than univariate TSC. The standard TSC benchmark, DTW, is still hard to beat and competitive with more recently proposed alternatives. HIVE-COTE, with components all built independently on each dimension, is significantly better than DTW. However, our multivariate DTW currently use full window DTW. Experience with the univariate case indicates that tuning the warping window will result in significant improvement. Furthermore, the data is not normalised by default. We found normalisation had no effect on HIVE-COTE, but it could considerably effect DTW.

The most promising algorithm that uses information between dimensions is MUSE. It performs as well as HIVE-COTE and is significantly better than DTW. However, MUSE carries a massive memory overhead which grows dramatically with problem dimensions. Even on the 21 problems that required less than 500 GB of memory, it averaged 26GB as opposed to HIVE-COTE, which averaged 1.6 GB.

Both shapelet algorithms performed relatively well, despite neither of them utlising information between dimensions. We speculate that shapelets are useful in dealing with long (EigenWorms) or high dimensional data (PEMS).

The deep learning algorithms have been disappointing in these experiments. Both TapNet and MLCN are very much in the middle of the pack. They have a tendency to occasionally completely fail. No doubt these will be improved over time for MTSC, but, as yet, they are not consistently state-of-the-art.

The UEA MSTC archive is fairly new and needs more development. New data are being added and donations are always welcome. An expanded version with at least 50 data sets is planned for 2020. These experiments represent a platform for future development. We would expect that, in the near future, algorithms that explicitly model interactions between dimensions would outperform all of the algorithms presented here and advance the research field of MTSC.

\end{document}